\documentclass[10pt,twocolumn,letterpaper]{article}

\usepackage{booktabs,multirow,adjustbox,diagbox,threeparttable,dsfont}
\usepackage[pagenumbers]{cvpr} 

\usepackage[dvipsnames]{xcolor}

\definecolor{cvprblue}{rgb}{0.21,0.49,0.74}
\usepackage[pagebackref,breaklinks,colorlinks,citecolor=cvprblue]{hyperref}
\newcommand\method{\texttt{BerfScene}\xspace}
\newcommand\pe{\texttt{pe}\xspace}

\usepackage[symbol]{footmisc}

\definecolor{textgreen}{RGB}{88, 133, 58}
\definecolor{textred}{RGB}{197, 90, 17}
\title{BerfScene: Bev-conditioned Equivariant Radiance Fields \\ for Infinite 3D Scene Generation}

\author{Qihang Zhang$^{1}$ \quad Yinghao Xu$^{2}$ \quad Yujun Shen$^{3}$ \quad Bo Dai$^{4}$ \quad Bolei Zhou$^{5\dagger}$  \quad Ceyuan Yang$^{4\dagger}$ \\[5pt]
	{$^1$CUHK \quad
	$^2$Stanford \quad
	$^3$Ant Group \quad 
	$^4$Shanghai AI Laboratory \quad 
    $^5$UCLA \quad }
}

\begin{document}

\maketitle

\footnotetext{\textsuperscript{$\dagger$} Corresponding authors}
\begin{abstract}

Generating large-scale 3D scenes cannot simply apply existing 3D object synthesis technique since 3D scenes usually hold complex spatial configurations and consist of a number of objects at varying scales.
We thus propose a practical and efficient 3D representation that incorporates an equivariant radiance field with the guidance of a bird's-eye view (BEV) map.
Concretely, objects of synthesized 3D scenes could be easily manipulated through steering the corresponding BEV maps. 
Moreover, by adequately incorporating positional encoding and low-pass filters into the generator, the representation becomes equivariant to the given BEV map.
Such equivariance allows us to produce large-scale, even infinite-scale, 3D scenes via synthesizing local scenes and then stitching them with smooth consistency.
Extensive experiments on 3D scene datasets demonstrate the effectiveness of our approach.
Our project website is at: \url{https://zqh0253.github.io/BerfScene/}.

\end{abstract}
\section{Introduction}\label{sec:intro}

The advancement in implicit and explicit 3D representations has driven the rapid progress in high-quality 3D object generation~\cite{graf, volumegan, skorokhodov2022epigraf, eg3d, stylenerf, stylesdf, hologan, gao2022get3d}.
However, directly applying object synthesis methods to 3D scene generation poses challenges due to inherent variations in spatial scales and composited objects within 3D scenes.
Considering that urban architects construct city scenes, they won't place building randomly but always starts from a detailed map, serving as a foundational guide outlining the spatial configurations of blocks and buildings.
This highlights the need for a suitable representation tailored for 3D scenes,  capable of streamlining the scene generation process.

A well-structured scene representation must capture spatial relationships between objects and have the flexibility to scale up, facilitating the generation of scenes on a large or infinite scale. 
Previous approaches often relied on scene graphs~\cite{ImageFromSceneGraph, sowizral2000scene, neuralscene, cunningham2001lessons} for representation, containing rich object relations but facing limitations in processing due to unstructured topology.
Recent work DiscoScene~\cite{xu2022discoscene} proposes representing scenes with a set of 3D bounding boxes. 
However, despite offering a volumetric depiction of objects, it introduces complexity in interpreting the entire scene and faces scalability challenges.

To overcome this, we choose a 2D bird's-eye-view (BEV) map to describe the scene structure, providing a practical and efficient way to represent and analyze spatial information, thereby simplifying the scene generation process.
Concretely, BEV map could specify the composition and scales of objects clearly. Generating a large-scale scenes could be thus formulated as synthesizing local scenes first and then composing them together. 
However, composing the local blocks into a coherent global scene in 3D space always leads to the severe artifacts such as jittering and inconsistency, as BEV maps can be ambiguous to fine-grained semantics \emph{i.e.}, primarily deliver a global layout and locations of objects but lack insights into the detailed visual appearance of the objects.

 \begin{figure}[t]
    \centering
        \includegraphics[width=\linewidth]{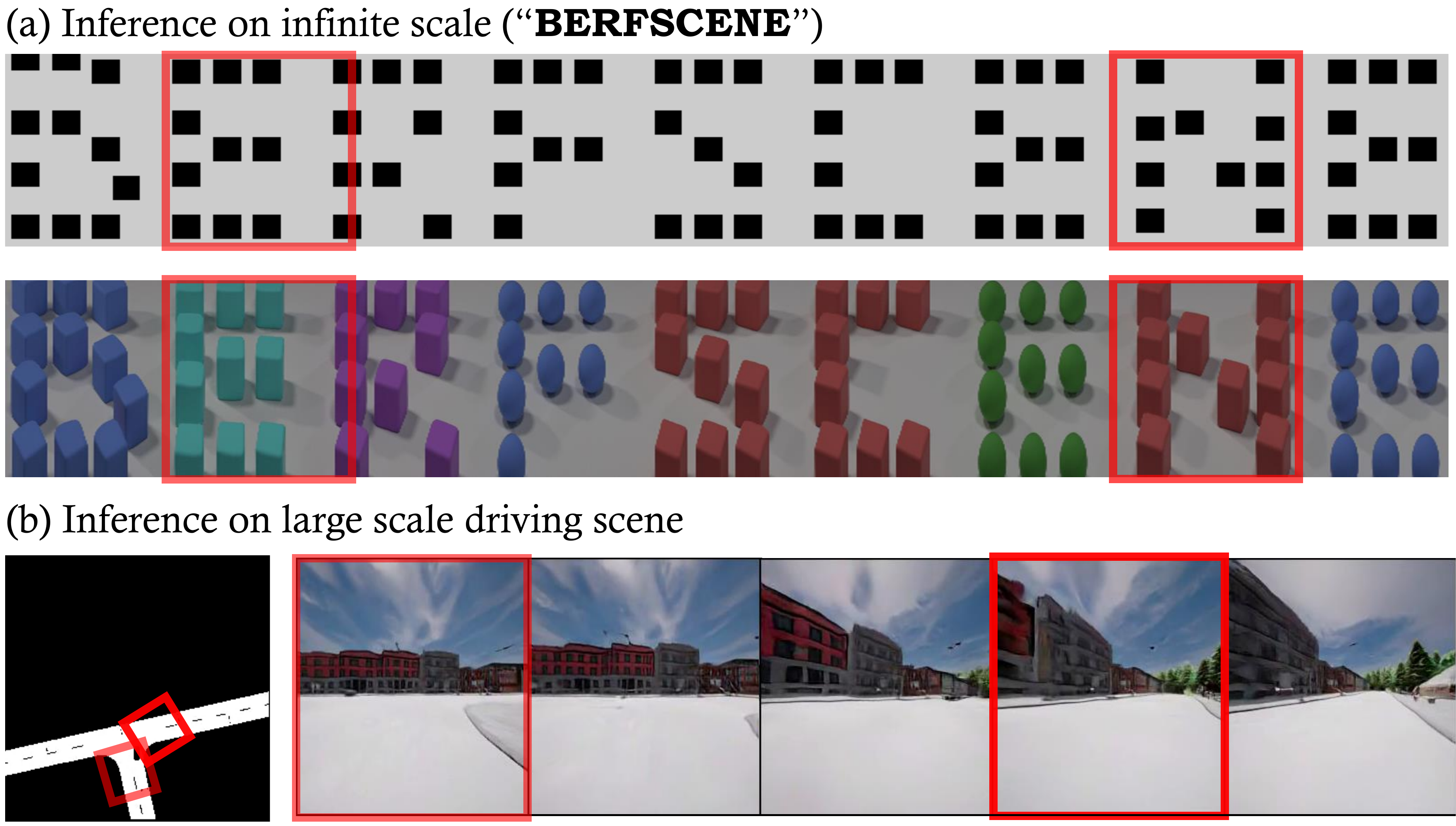}
        \vspace{-10pt}
    \caption{
       \textbf{\method focuses on unbounded 3D scene synthesis.} Above: a CLEVR scene labeled "BERFSCENE". Below: a driving scenario before and after executing a right turn. }  
    \label{fig:teaser}
    \vspace{-15pt}
\end{figure}

To avoid the ambiguity of BEV maps, recent attempts like InfiniCity~\cite{lin2023infinicity} and SceneDreamer~\cite{scenedreamer} incorporate explicit 3D structures (e.g., voxels) as a hard constraint to ensure the continuity of the composition process.
However, collecting and loading large-scale 3D structures always pose the computational overhead and inefficiency.
Alternatively, we tackle this issue by integrating the equivariance property with a carefully-designed architecture into the BEV-conditioned representation. The consistency across various local scene generation is accordingly enhanced. 

By introducing BEV-conditioned Equivariant Radiance Fields based on such representation, we present \method, a framework allowing for large-scale 3D scene synthesis and flexible editing of camera pose and composite objects, as shown in \cref{fig:teaser}. 
Our generator is conditioned on a local BEV patch to learn the entire scene's distribution, utilizing a specific network architecture to maintain equivariance across the same semantic regions in different BEV maps. 
This design incorporates extra padding and low-pass filters~\cite{stylegan3, zhang2019making} in the generator to reduce aliasing, ensuring consistent synthesis reflecting specified spatial configurations under any local coordinates. 
Thanks to the equivariance of the BEV-conditioned representations, our method can learn from 2D images showing scenes with limited spatial extent, while also being capable of generating infinite-scale 3D scenes.
We evaluate our method on 3D scene datasets including CLEVR~\cite{clevr}, 3D-Front~\cite{3dfront,3dfuture}, and Carla~\cite{dosovitskiy2017carla}. Through qualitative and quantitative experiments, we demonstrate that our method achieves state-of-the-art performance in generating large-scale 3D scenes.
\section{Related work}\label{sec:relatedwork}

\noindent\textbf{3D-aware image synthesis.}
We have witnessed amazing progress in image generation with 2D GANs~\cite{gan, pggan, stylegan, stylegan2, stylegan3}.
Recent works lift 2D GANs with 3D inductive bias for 3D-aware generation from unstructured single-view image collections.
Early works leverage the voxel~\cite{von, hologan, schwarz2022voxgraf}, mesh~\cite{gao2022get3d,xiong2023get3dhuman}, depth~\cite{shi20223d} or 2D feature plane~\cite{skorokhodov2022epigraf, skorokhodov20233d} to explicitly model object structure, but suffer from poor visual fidelity and geometric consistency.
Another line of research integrates the neural radiance fields~\cite{graf,pigan,gram, shadegan, gof, geod, d3d} into the GAN generator to alleviate these limitations.
Recently, diffusion models have been used to synthesize 3D-aware images by distilling knowledge from large pretrained text-to-image models~\cite{poole2022dreamfusion} or by training from scratch with direct 3D supervision~\cite{pointe, shape, tridiff,  3dgen, auto3d} or adopt image-to-image translation framework using view conditioning~\citep{zero123, genvs,nerfdiff,instant3d, dmv3d2, one2345, syncdreamer}.
 However, these methods primarily focus on object modeling and have limited capacities in generating large-scale scenes, which our method focuses on.

\noindent\textbf{3D scene generation.} 
Although 3D aware-image and object generation has been significantly advanced in recent years, 3D scene generation remains a challenging task since generating a 3D scene usually considers the composition of objects and their corresponding scales. 
To tackle these issues, recent attempts explore to leverage existing prior like layout~\cite{nguyen2020blockgan, giraffe, xue2022giraffehd, xu2022discoscene}, grid plane~\cite{gsn,kim2023neuralfield}, depth maps~\cite{shi2022deep, skorokhodov20233d, shen2022sgam}, or voxels~\cite{scenedreamer, lin2023infinicity}  to generate 3D scenes. 
We follow this philosophy yet incorporate a BEV-map as the conditions as it enables the flexible division of a large-scale scenes \emph{i.e.}, specifies the scene configuration clearly. 
A very related work CC3D~\cite{bahmani2023cc3d} shares similarities with our approach in utilizing a bird's-eye-view (BEV) map as a conditioned layout for generating scene radiance fields. However, CC3D is limited in its ability to generate infinite 3D scenes due to its lack of composition modeling. In contrast, our model overcomes this limitation by employing an equivariant representation conditioned on BEV maps, enabling seamless composition and facilitating the generation of infinite-scale scenes.

 \begin{figure*}[t]
    \centering
        \includegraphics[width=\textwidth]{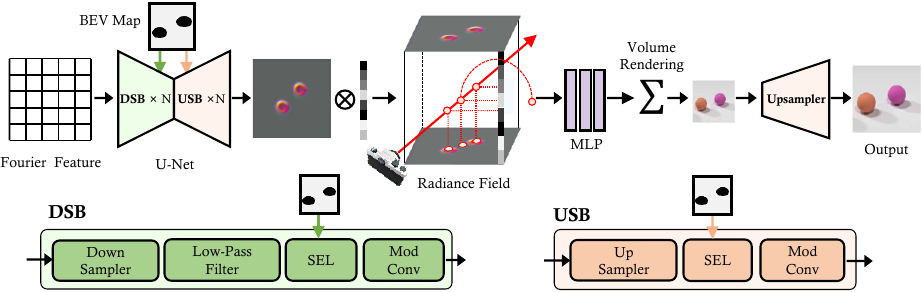}
    \caption{
        \textbf{Illustration of \method:} 
        A U-Net takes the fourier feature as input and gradually down-samples (DSB) and up-samples (USB) the features. The internal features would be spatially modulated by the BEV maps via SEL block, resulting in a BEV-conditioned radiance field. With the anti-aliasing design (\emph{e.g.}, low-pass filters), the entire synthesis pipeline becomes equivariant to the BEV maps. 
            } 
    \label{fig:pipeline}
    \vspace{-10pt}
\end{figure*}

\section{Method}\label{sec:method}

\method employs a BEV map as an input to specify a scene and generates a radiance field conditioned on the BEV representation, which is then used for image synthesis through volume rendering. 
To support large-scale scene generation, the BEV-conditioned radiance field is further extended into an equivariant representation through a carefully designed feature extractor.
We first introduce preliminary knowledge about volume rendering in ~\cref{method:pre}.
In ~\cref{method:representation}, we discuss the design of the equivariant representation.
~\cref{method:framework} describes the scene generation framework, including implementation, training, and inference details.

\subsection{Preliminaries}\label{method:pre}
 
The neural radiance field~\cite{nerf} has gained tremendous popularity among recent works in view synthesis and image generation.
Specifically, to render an image given a camera viewpoint, multiple rays are cast out, with $N$ points $\{p_i | i=1,\cdots,N\}$ sampled along each ray $r$. 
For each point $p_i=(x_i, y_i, z_i)$, we query its color $c_i$ and density $\sigma_i$:
\begin{equation}
    \label{eq: ray query}
    c_i, \sigma_i = \Theta(f(p_i), d),
\end{equation}
where $f(p_i)$ is the encoding feature of $p_i$, $d$ is the ray direction, and $\Theta$ is parameterized as a Multi-Layer Perceptron (MLP).
The color of the ray $C(r)$ is further calculated as the weighted average of each point's color:
\begin{align}
    C(r) &= \sum_{i=1}^N \left[ \prod_{j=1}^{i}e^{(-\sigma_j\delta_j)}\cdot (1-e^{(-\sigma_i\delta_i)}) \right] c_i, 
\end{align}
where $\delta_i$ is the length of the $i$-th interval on the ray.

As for  $f(\cdot)$ , there are different design choices to encode each single point, like positional embedding: $f(p_i)=(\pe(x_i), \pe(y_i), \pe(z_i))$, and sampled feature from 2D feature map: $f(p_i)=(\Phi(U_{xy}, x_i, y_i), \Phi(U_{xz}, x_i, z_i), \Phi(U_{yz}, y_i, z_i))$, where $U_{xy},U_{xz},U_{yz}$ denote learnable 2D feature map and $\Phi$ denotes feature sampling operation.

Recent works additionally sample latent code $s$ and incorporate it into the encoding feature $f(\cdot)$ for 3D-aware image generation~\cite{graf, pigan, gram}. 
For example, EG3D~\cite{eg3d} encodes each point feature as: $f(p_i)=(\Phi(U_{xy}(s), x_i, y_i), \Phi(U_{xz}(s), x_i, z_i), \Phi(U_{yz}(s), y_i, z_i))$, where $U_{xy}(s),U_{xz}(s),U_{yz}(s)$ are generated 2D feature map conditioned on latent code $s$. 
Our work follows this line of works.

\subsection{Equivariant BEV-conditioned representation for radiance field}\label{method:representation}

Given that various scenes (\emph{e.g.}, traffic scenes) could be represented by a ground plan, we propose to leverage Bird-Eye-View (BEV) map to steer the generation of the radiance field. We also improve the equivariance of the representation for large-scale scene synthesis. In this section, we will provide a detailed explanation of our design.

\noindent\textbf{BEV-conditioned radiance field.} 
In order to incorporate the prior information provided by the BEV map, we introduce a generator $U$ that generates a conditioned 2D feature map. 
The network architecture of $U$ is a U-Net architecture with StyleGAN blocks. 
As illustrated in \cref{fig:pipeline}, the generator takes a 2D Fourier feature map $\gamma$ as input and progressively modifies the feature map using sequential encoders and decoders, which are modulated by a randomly sampled latent code $s$ and the BEV map $\mathcal{B}$.

We incorporate the 2D Fourier feature map $\gamma$ as the input to provide positional information for local radiance field. It is defined on a positional grid $\mathbf{v}$ that spans the global coordinates. Each position in the grid is associated with a specific value.:
\begin{equation}
\begin{aligned}
    \gamma(\mathbf{v})=&\left[a_1 \cos \left(2 \pi \mathbf{b}_1^{\mathrm{T}} \mathbf{v}\right), a_1 \sin \left(2 \pi \mathbf{b}_1^{\mathrm{T}} \mathbf{v}\right), \ldots,\right. \\
    &\left. a_m \cos \left(2 \pi \mathbf{b}_m^{\mathrm{T}} \mathbf{v}\right), a_m \sin \left(2 \pi \mathbf{b}_m^{\mathrm{T}} \mathbf{v}\right)\right]^{\mathrm{T}},
\end{aligned}
\end{equation}
where $a_i, \mathbf{b}_i$ denotes predefined amplitudes and Fourier basis frequencies. 
Each subsequent encoder or decoder uses Spatial Encoding Layer (SEL)~\cite{wang2022improving} to incorporate the BEV map $\mathcal{B}$. Concretely, given an intermediate feature map $a$, one block in U-Net operates as:
\begin{align}
    a' &= \texttt{SEL}(\mathcal{T}(a), \mathcal{T}(E(\mathcal{B}))), \\
    a'' &= \texttt{ModConv}(a', s),
\end{align}
where $E$ is an encoder with two convolutional layers that extracts BEV's feature map,  $\mathcal{T}(\cdot)$ denotes the interpolation operation that resizes the two feature maps, and \texttt{ModConv} performs the modulated convolution~\cite{stylegan2} to further modify the features based on the latent code $s$.

The output feature map of the U-Net $U$ is then lifted to 3D by computing cross product with the positional embedding of coordinate $z$: $U(\mathcal{B}, \gamma, s)\times \{\text{pe}(Z)\}$.
Consequently, color and density can be obtained via:
\begin{equation}
   c,\sigma = \Theta(\Phi(U(\mathcal{B}, \gamma, s)\times \{\text{pe}(Z)\}, x, y, z)), \\
    \label{eq: feature sampling}
\end{equation}
where $\Phi$ is the feature sampling operation, and $\Theta$ is the MLP that takes sampled features as the input.

\noindent\textbf{Equivariant property.}
Building upon previous designs, we have the capability to create scenes according to the BEV map. 
As a result, this allows us to synthesize a scene of infinite scale in a divide-and-conquer fashion, \emph{i.e.}, dividing a global map into local patches, generating local scenes, and composing them together.
However, the BEV conditioned radiance field can result in uncertainty in terms of fine-grained details.
This uncertainty may lead to the synthesis inconsistency since the same objects may appear in multiple local scenes, substantially deteriorating the quality when composing several local scenes for large-scale scene synthesis. 
We thus seek for the guarantee of the equivariant property.

In particular, regular convolutions with padding and down-sampling tend to cause aliasing~\cite{stylegan3, zhang2019making}, \emph{i.e.}, the synthesized concepts are strongly related to their coordinates. 
Considering this, we carefully design the operations in U-Net to maintain the equivariance to a maximum extent.
1) BEV with a wide margin: as border padding would leak the absolute positional information to the internal representations~\cite{islam2020much, kayhan2020translation, xu2021positional}, we follow \cite{stylegan3} to leave a large margin around BEV map to ensure the representation unimpeded by padding. 
2) Low-pass filters: it is inevitable to down-sample the internal feature maps, for the sake of memory efficiency. 
According to Nyquist Law~\cite{shannon1949communication}, the representation capacity of a regularly sampled signal is bound by half of the sampling rate. Otherwise, excessive signals can cause aliasing.
Therefore, we introduce the low-pass filter before down-sampling to restrict the representation within a reliable region. 
The transform for the downsample becomes
\begin{align}
\mathcal{T}(\cdot) = \texttt{Low-Pass}(\cdot)\circ \texttt{Interp}(\cdot), 
\end{align}
where the low pass filter is designed as a finite impulse response (FIR) filter.
With this equivariant property, generating large scale scenes becomes simply composing multiple local scenes, with the consistent concepts.

\subsection{Scene generation framework}\label{method:framework}

With the equivariant BEV-conditioned representation designed above, we now introduce \method, the proposed method for infinite 3D scene generation.

\noindent\textbf{Generator.}
The generator consists of a U-Net encoder that produces the spatial feature map for the volume rendering in~\cref{eq: feature sampling}. 
Concretely, this U-Net encoder takes Fourier feature as input, where the internal features would be modulated via the latent code.
Besides, BEV map $\mathcal{B}$ would be incorporated into this encoder through the SEL, which could further guide the spatial configurations of the final synthesis. 
As internal feature maps are gradually down-sampled, we apply the low-pass filters to remove the excessive frequencies, improving the equivariance of theis encoder. 
To this end, the output feature map of this unet manages to correctly and equivariantly reflect the spatial structure determined by BEV maps.
Given this feature maps, images would be obtained through the neural rendering.

\noindent\textbf{Discriminator.} We follow the dual-discriminator design of EG3D~\cite{eg3d}. A bi-linearly upsampled version of the rendered image is concatenated with the super-resolved version. The discriminator takes as input the resulted six-channel image. 

\noindent\textbf{Training objectives.}
During training, style code $s$ is randomly sampled from Gaussian distribution.
BEV map $\mathcal{B}$ and camera pose $\gamma$ are randomly sampled from the dataset.
We optimize traditional adversarial loss $\mathcal{L}_{adv}$, $R_1$ regularization loss $\mathcal{L}_{R_1}$, and density regularization loss $\mathcal{L}_{density}$ as proposed in \cite{eg3d}.
The overall training target is a weighted sum of the above loss terms:
\begin{equation}
    \mathcal{L} = \lambda_{adv}\mathcal{L}_{adv} +  \lambda_{R_1}\mathcal{L}_{R_1} +  \lambda_{density}\mathcal{L}_{density},
    \label{eq: loss}
\end{equation}
where $\lambda_{adv}, \lambda_{R_1}, \lambda_{density}$ are weighted coefficients.

\noindent\textbf{Inference of infinite-scale synthesis.}
Rather than generating high-quality local scene images, \method supports scene generation at an infinite scale.
After defining a global BEV map, we divide it into several local BEVs and render images conditioned on them.
One can get a progressively moving video by continuously cropping local BEVs.
In addition, broad-view image can be generated by stitching rendered results.
We also adopt supersampling anti-aliasing (SSAA) to perform ray marching at a temporary higher resolution and downsample the feature map to the original resolution.
SSAA suppresses aliasing effect and provides better visual quality.

\section{Experiments}\label{sec:exp}

We evaluate \method on diverse datasets, and compare it with baseline methods of both image generation and 3D-aware image generation.

\subsection{Settings}

\noindent\textbf{Datasets.} We conduct experiments on three datasets: CLEVR~\cite{clevr}, 3D-Front~\cite{3dfront, 3dfuture}, and Carla~\cite{dosovitskiy2017carla}.
CLEVR is a multi-object dataset with a 3D rendering engine.
We use the official script to render images for training and evaluation.
The camera position is fixed in the global coordinate.
For each scene, we randomly place 3 to 8 objects with various colors and shapes.
We collect 80,000 images in $256\times256$ resolution.
3D-Front is a 3D indoor scene dataset with diverse furniture including bed, wardrobe, \textit{etc}.
We randomly place the camera and collect 50,000 images in $256\times 256$ resolution on this dataset, covering 2535 different scenes in total. 
Carla is a driving simulator with realistic visual appearance. It covers different weather conditions, and diverse road environments (from rural to urban). We collect 28,000 frames in $256\times 256$ resolution.

\noindent\textbf{Metrics.}
Following the prior, we use the Frechet Inception Distance (FID) as a quantitative metric to evaluate the quality of our image synthesis results. We sample 50K real images and 50K generated samples to compute the FID score.
Additionally, we measure the consistency of the same scene under different local coordinates to test composition feasibility. 
Since it is challenging to directly compare generated 3D scenes, we approximate the scene using the rendered image $G(\mathcal{B}, z)$.
Following \cite{stylegan3}, we report the peak signal-to-noise ratio (PSNR) in decibels (dB) between two sets of images obtained by translating the input and output by a random amount:
\begin{equation}
\begin{aligned}
    \text{EQT} &= 10\cdot\text{log}(\frac{I_{max}^2}{\mathds{E}_{s, x}(\|G(t_x[\mathcal{B}], s) - t_x[G(\mathcal{B}, s)]\|)}), 
\end{aligned}
\end{equation}
where $t_x[\cdot]$ stands for translation operation by $x$ margin, and the intended dynamic range of generated images from $-1$ to $+1$ gives $I_{max}=2$.

\noindent\textbf{Baseline.}
We compare our method to both 2D and 3D GANs. Specifically, we evaluate our approach against StyleGAN2, EG3D, and GSN to explore the impacts of introducing inductive biases, such as equivariance, on image quality. Additionally, we assess our capacity of model for generating compositional 3D scenes using CC3D, which is a scene generation framework conditioned on BEV. 

\noindent\textbf{Implementation details.}
We follow the architecture design of EG3D except our equivariant BEV-conditioned generator.
To determine the best $R_1$ regularization weight, we performed a grid search across various datasets and methods. 
The values of $R_1$ regularization weight used in our experiments are available in the supplementary material. 
All other hyper-parameters were kept the same as EG3D. 
We conducted all experiments on 8$\times$A100 GPUs with a batch size of 64. More details can be found in the supplementary material.

\begin{figure*}[t]
    \centering
    \includegraphics[width=0.98\textwidth]{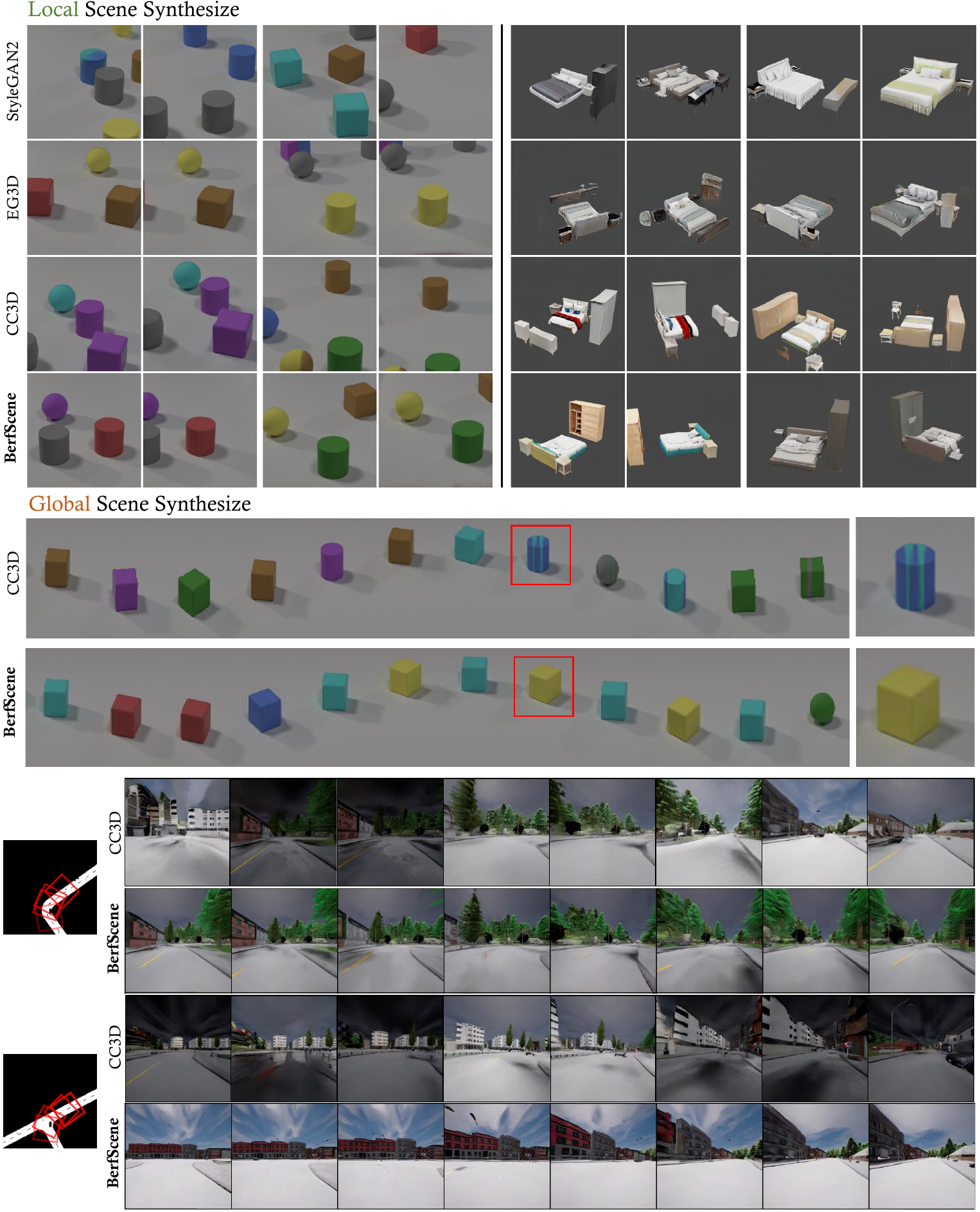}
    \caption{
        \textbf{Qualitative results of \textcolor{textgreen}{Local} Scene Synthesize} in $256\times256$ resolution on various datasets, and \textbf{\textcolor{textred}{Global} Scene Synthesis} on CLEVR. } 
    \label{fig:comparison}
\end{figure*}

\begin{figure*}[t]
    \centering
    \includegraphics[width=1.0\textwidth]{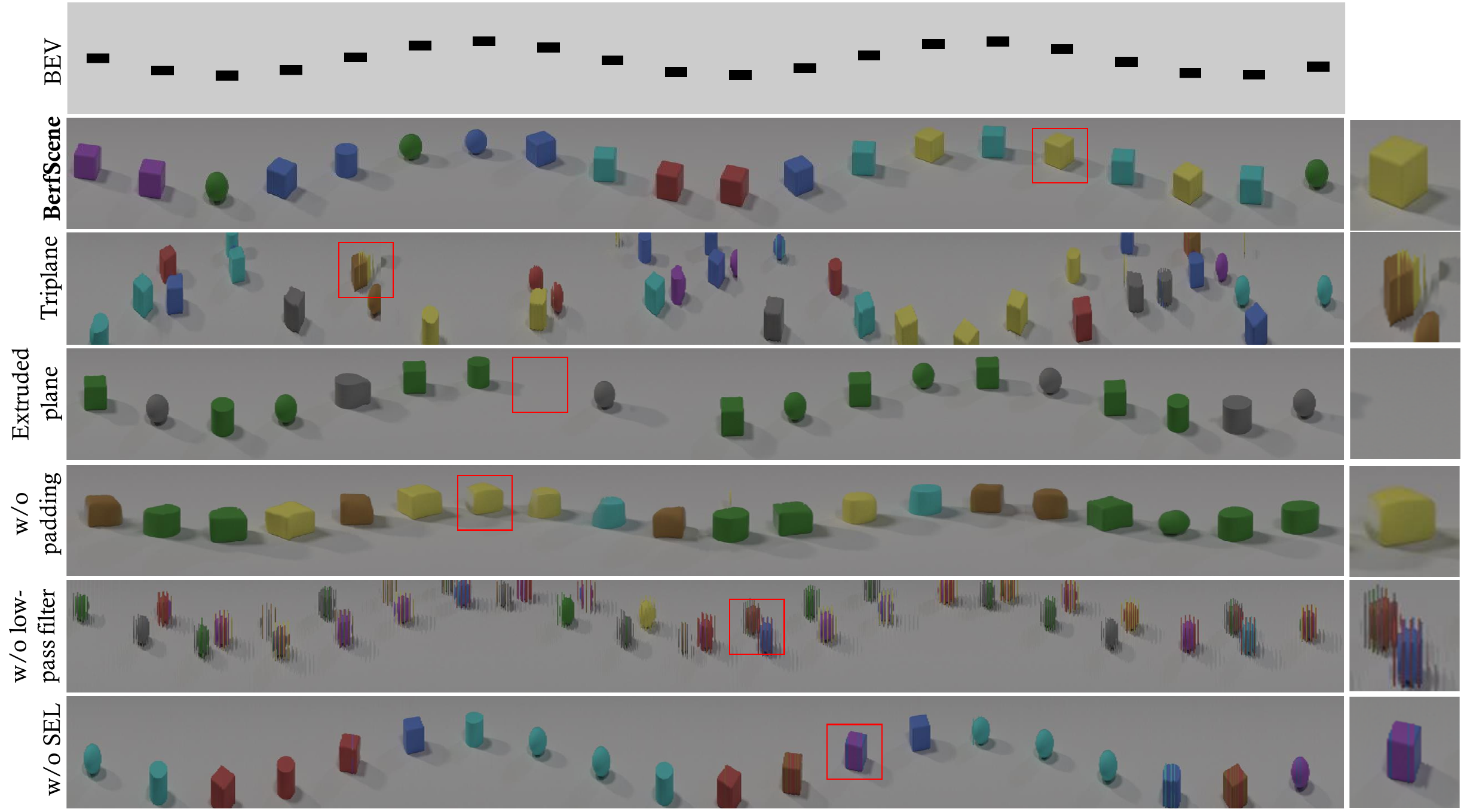}
    \caption{
       \textbf{Qualitative comparison for ablations} on large-scale scene synthesis. }
    \label{fig:large clevr}
    \vspace{-10pt}
\end{figure*}

\begin{table}[t]
\centering\footnotesize
\setlength{\tabcolsep}{7pt}
\caption{\textbf{Evaluation with baselines.} FID and EQT are reported as evaluation metrics. Note that we highlight the best results among 3D-aware models.}
\label{mainexp}
\vspace{-10pt}
\begin{tabular}{@{}llllll@{}}
\toprule
\multirow{2}{*}{Method} & \multicolumn{2}{c}{CLEVR} & \multicolumn{2}{c}{Front-3D} & \multicolumn{1}{c}{Carla} \\ \cmidrule(l){2-6} 
                        & FID (\textdownarrow)        & EQT    (\textuparrow)       & FID   (\textdownarrow)          & EQT (\textuparrow)      & FID   (\textdownarrow)            \\ \midrule
\textit{StyleGAN2}            &      6.95 & - & 31.01 & -  & 16.89                                                    \\ \midrule
GSN & -&- & 130.70 & - & - \\
EG3D      & 4.67 & - & 80.70 & - & 46.8 \\ 
CC3D                   &  3.61           &   21.94          &     42.88        &    14.74  & 45.2    \\ 
\midrule
Ours                    &    \textbf{0.96}      &    \textbf{22.02}           &   \textbf{36.78}          &      \textbf{15.76}      & \textbf{40.7}    \\ \bottomrule
\end{tabular}
\vspace{-10pt}
\end{table}

\subsection{Generation Results}

\noindent\textbf{Qualitative results.}
In \cref{fig:comparison}, we present results of local and global scene synthesis from our method and the baselines. 
For local scene synthesis, StyleGAN2, as a 2D image generator, cannot support the explicit camera control. On the contrary, we show two different views of one single scene for EG3D, CC3D, and ours.

When tested on the CLEVR dataset, StyleGAN2 fails to generate consistent object appearances. 
In the first example of StyleGAN2, the generated cylinder has a mixed color that is not present in the dataset. 
Both EG3D and CC3D suffer from blurry results, with slight blurs found in the generated output and twisted edges can be seen in CC3D's  results. 
In contrast, our method consistently produces high-fidelity images and also supports excellent camera control, as evidenced by the consistent results across different camera angles.
On the 3D-Front dataset, StyleGAN2 generates indoor scenes with high fidelity. 
EG3D fails to generate consistent results as the texture and shape vary across different camera poses. 
CC3D generates inaccurate shapes for small objects like nightstands and is leaning to generate blurry textures. 
In contrast, our method can generate indoor furniture with decent and consistent appearance across different camera views, demonstrating the effectiveness of our proposed scene representation.

Since both CC3D and our method are conditioned on BEV maps, we can continuously roll out BEV patches and generate and compose local scenes for global scene synthesis. 
We test the capacity of large scene synthesis on CLEVR and Carla.
For CLEVR, CC3D generates transient color of a single object and blurry edges, indicating fractional shaking across local patches. 
Our method can generate a high-fidelity global scene without any inconsistencies or blurs.
For Carla, our method can generate high quality driving videos with consistent visual appearance and 3D geometry of buildings and trees. Yet CC3D produces flickering frames with severe inconsistency.

\noindent\textbf{Quantitative evaluations.}
\cref{mainexp} reports the quantitative results (FID and EQT) over different methods\footnote{We failed to train GSN on CLEVR with \href{https://apple.github.io/ml-gsn/}{the official implementation}, hence we do not report the quantitative results.}.
On CLEVR, \method achieves a FID score of 0.96, a far better result compared to other methods.
Regarding 3D-Front, our method also gains a significant lead among all 3D GANs.
Additionally, our method consistently outperforms other 3D GANs in terms of EQT, demonstrating that our approach not only generates realistic 3D scene images but also enjoys good equivariance. This property is essential for composing local scenes into a large-scale scene, making our method a promising solution for generating 3D scenes of arbitrary scales.

\subsection{Ablation Study}
To better understand the individual contributions, we ablate main components by comparing quantitative metrics and qualitative large-scale scene synthesis results. 

\noindent\textbf{Radiance field representation design.}
To guide the generation process using BEV maps, we incorporated the Spatial Encoding Layer (SEL) into our generator to fuse the BEV. The output BEV feature map is further extended by positional embedding over the coordinate $z$ to create the radiance field representation. We compare this design to the triplane representation and 2D-to-3D extrusion proposed by \cite{bahmani2023cc3d}. To ensure a fair comparison, all designs share the same backbone, with the last convolutional layer having different output channels. 
Our design output 32 channels, while the triplane representation triples the channel number, and the 2D-to-3D extrusion produces $32\times N$ channels, where $N$ is the number of height dimension channels. 
In \cref{exp: repre design}, worse performance on FID and EQT is observed for both triplane and extruded plane designs.
In \cref{fig:large clevr},  the generated global scenes with these two designs also suffer from severe artifacts.

\begin{table}[t]
\centering\footnotesize
\setlength{\tabcolsep}{8pt}
\caption{\textbf{Ablation study over different backbone design choices.}}
\label{exp: repre design}
\vspace{-10pt}
\begin{tabular}{@{}llllll@{}}
\toprule
 \multirow{2}{*}{Configuration}               & \multicolumn{2}{c}{CLEVR} & \multicolumn{2}{c}{Front-3D} \\ \cmidrule(l){2-5} 
                                                     & FID (\textdownarrow)        & EQT    (\textuparrow)       & FID   (\textdownarrow)          & EQT (\textuparrow)           \\ \midrule
Triplane                      &          18.11            &   19.58            &    39.17           &     14.10         \\
Extruded plane                                        &     5.60        &   20.13          &    50.40          &     15.29         \\ \midrule
Ours  &                           \textbf{0.96}      &   \textbf{22.02}          &      \textbf{36.78}        &    \textbf{15.76}    \\ \bottomrule
\end{tabular}
\vspace{-10pt}
\end{table}

\begin{table}[h]
\centering\footnotesize
\setlength{\tabcolsep}{8pt}
\caption{\textbf{Ablation study over design components.}}
\label{exp: ablation}
\vspace{-10pt}
\begin{tabular}{@{}lllll@{}}
\toprule
 \multirow{2}{*}{Configuration}          & \multicolumn{2}{c}{CLEVR} & \multicolumn{2}{c}{Front-3D} \\ \cmidrule(l){2-5} 
                     & FID (\textdownarrow)        & EQT    (\textuparrow)       & FID   (\textdownarrow)          & EQT (\textuparrow)         \\ \midrule
w/o padding BEV      &   2.50        &    19.01        &   51.30          &  13.32       \\
w/o low-pass filter  &   5.53         &     18.19        &     36.87       &     14.45         \\ 
w/o SEL              &    6.27         &     22.00           &     45.90                &   15.41           \\          \midrule
Ours                 &  \textbf{0.96}        &   \textbf{22.02}          &    \textbf{36.78}       &    \textbf{15.76}           \\ \bottomrule
\end{tabular}
\vspace{-10pt}
\end{table}

\noindent\textbf{Padding BEV.}
To analyze how additional padding suppresses aliasing, we compare models trained on BEVs with and without padding.
As can be seen in  \cref{exp: ablation}, EQT drops by a large margin.
This result indicates that positional information leaks into the generator and disrupts the equivariance property, limiting models for large-scale scene generation (also see wierd shapes in \cref{fig:large clevr}).

\begin{figure*}[t]
    \centering
    \includegraphics[width=1.0\textwidth]{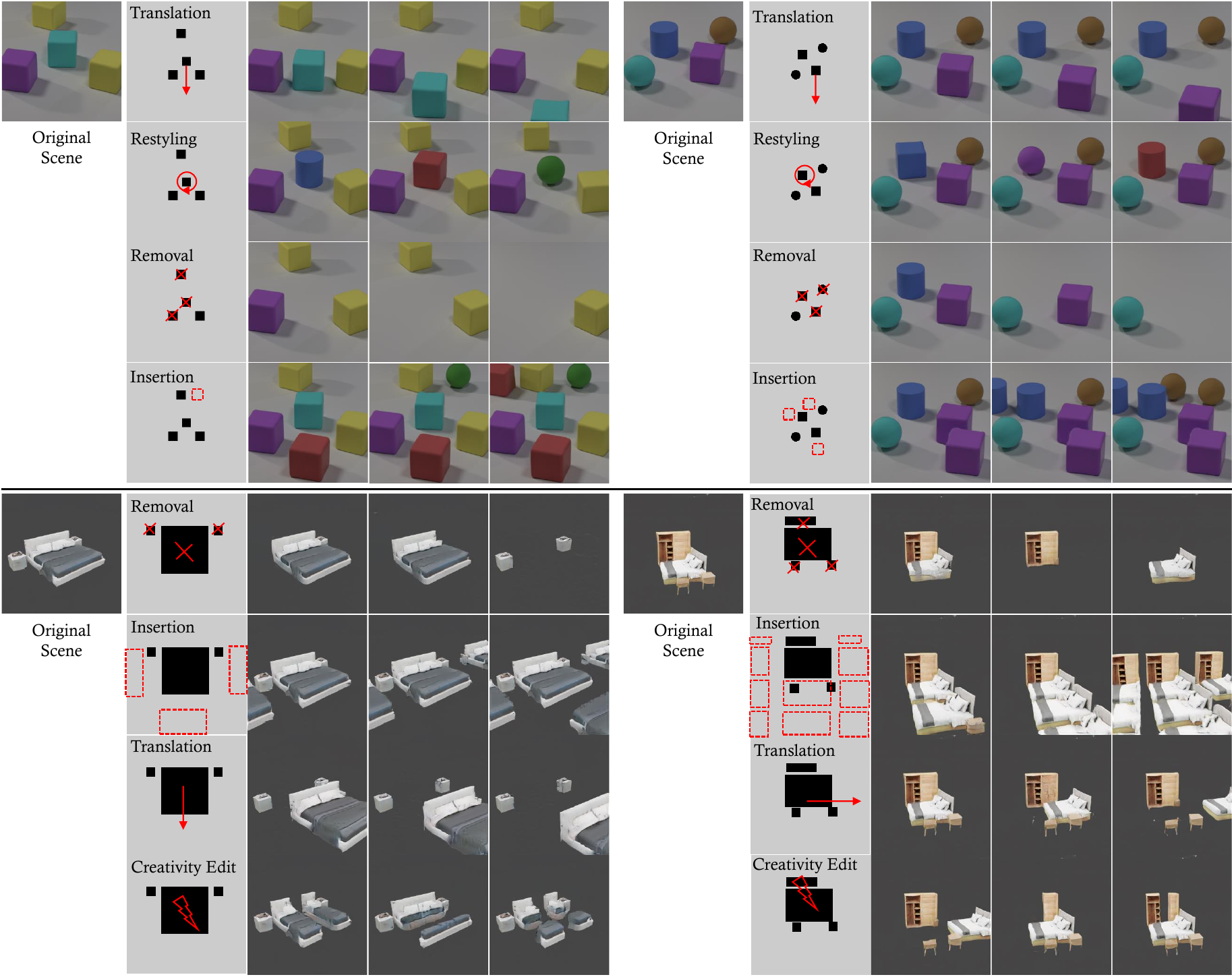}
    \vspace{-10pt}
    \caption{\textbf{Controllable 3D scene synthesis in $256\times256$ resolution.}  We perform versatile user control on the scene objects by varying BEV map, such as translation, restyling, removal, insertion.
    } 
    \label{fig:control}
    \vspace{-2pt}
\end{figure*}

\noindent\textbf{Low-pass filters.}
Beside padding in CNNs, aliasing could also be caused by excessive high frequency noise after down sampling layers.
We study whether low-pass filter helps alleviate it in our 3D generation scenario.
After removing low-pass filters in the network, EQT goes down, with intense discontinuity observed in the generated global scene, demonstrating that low-pass filters are essential to guarantee the equivariance property.

\noindent\textbf{SEL layer.}
In our U-Net backbone, BEV map is repeatedly fused into the feature map through SEL layer to achieve precise layout control.
An alternative choice is to directly feed BEV into the backbone.
As shown in \cref{exp: ablation}, FID increases by a large margin compared to our method with SEL.
We hypothesize that repeated SELs could make the best of the geometry guidance from BEV, and thus generates scenes with more realistic and relevant spatial configurations.

\section{Applications of BerfScene}

\subsection{Infinite scene generation}
Our method can generate large-scale, even infinite, scenes, by dividing a global scene into local patches, generating and then seamlessly composing them.
Concretely, we use sliding window to get continuous local BEV maps. These maps serve as the conditioning input for generating a \textit{navigating video}.
Then, we extract the middle vertical line from each frame in the video and stack them to form a holistic view of the entire scene.
We demonstrate generated large-scale scenes with various layouts in \cref{fig:large clevr}.

\subsection{Scene editing}

Our generator is conditioned on the BEV map, thus it is easy to edit the scene by varying the input BEV map. 
In \cref{fig:control}, we demonstrate different scene editing results including \textbf{1) translation}, a user can rearrange objects' layout; \textbf{2) restyling}, a user can directly modify single object's semantic to achieve restyling; \textbf{3) removal and insertion}, a user can delete or copy objects from the scene.

\section{Discussion}\label{sec:conclusion}

\noindent\textbf{Limitations.}
Although infinite-scale scene generation has been enabled, there remain several limitations we would like to discuss.
First, as we follow the generative radiance field that mainly learns from the training set, the view of camera for inference is quite limited for large-scale scene synthesis. Collecting data with more diverse observations may help alleviate this issue. 
Second, current design only supports the static scene generation. How to enable the large-scale dynamic scene synthesis remains open for future work.
Furthermore, it is important to note that our method may encounter challenges in achieving precise attribute control due to the absence of explicit supervision. For instance, when specifying a particular color in the BEV map, the synthesized output may exhibit a different color. This could potentially be enhanced by incorporating CLIP supervision.

\noindent\textbf{Conclusion.}
This work introduces \method that can generate 3D scenes of arbitrary scales. 
We propose a BEV-conditioned radiance field to represent a 3D scene. This approach enables users to directly steer the generated spatial configurations via BEV maps.
To ensure smooth and consistent composition of multiple scenes, we further ensure the equivariance of the BEV-conditioned representations. We introduce several architectural designs, including a wider margin and low-pass filters, to achieve this goal. As a result, we can synthesize infinite-scale scenes by simply composing multiple syntheses controlled by local BEV maps.
Experimental results on various 3D scene datasets demonstrate the effectiveness of our proposed method.

{\small
\bibliographystyle{ieeenat_fullname}
\bibliography{main}
}

\maketitlesupplementary
\appendix
\newcommand{\AppendixPrefix}{A}
\renewcommand{\thefigure}{\AppendixPrefix\arabic{figure}}
\setcounter{figure}{0}
\renewcommand{\thetable}{\AppendixPrefix\arabic{table}} 
\setcounter{table}{0}
\renewcommand{\theequation}{\AppendixPrefix\arabic{equation}} 
\setcounter{equation}{0}

\section{Datasets Details}

In this section, we introduce the datasets we use and show sampled BEV maps and front view images.

\begin{figure*}[t]
    \centering
    \includegraphics[width=1.0\textwidth]{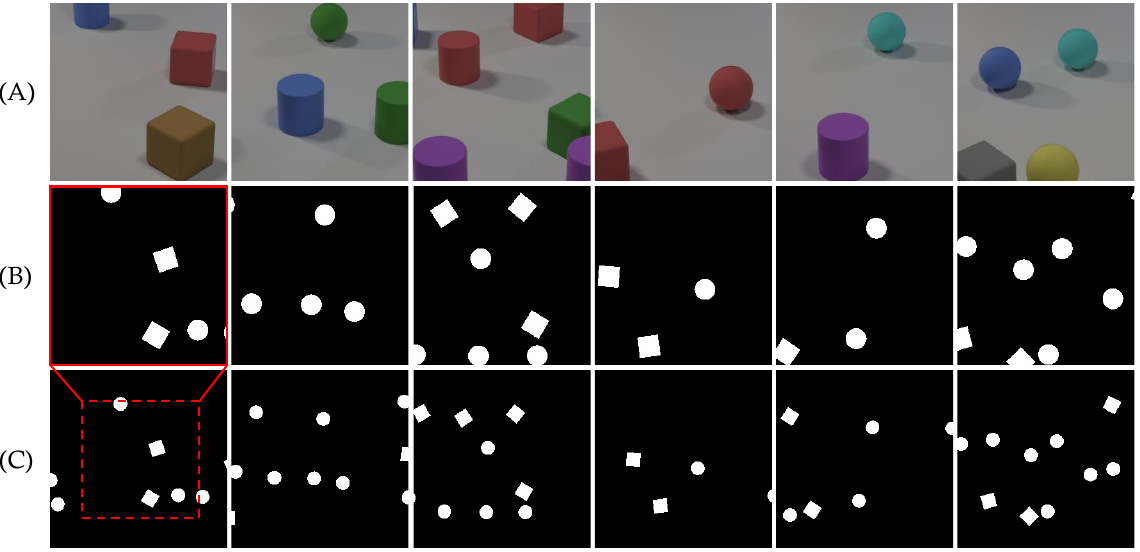}
    \caption{
        \textbf{Sampled front view images and BEV maps on CLEVR.} Row A shows rendered front view images. Row B and C show corresponding BEV maps without and with broader paddings.
        } 
    \label{fig:supp clevr samples}
\end{figure*}

\noindent\textbf{CLEVR.} CLEVR~\cite{clevr} is a synthetic dataset, containing cubes, spheres, and cylinders with different colors. 
We adopt the official script\footnote{https://github.com/facebookresearch/clevr-dataset-gen} for rendering. 
80K images are collected in total. The camera positions are fixed for all the images.
In \cref{fig:supp clevr samples}, we show rendered images with their corresponding BEV maps.
We demonstrate tight BEV maps used in the ablation study which represent just the right amount of objects as in the front views. In addition, we also show BEV maps with broader paddings for improving the equivariance of the BEV-conditioned representation.
We concatenate together a one-hot vector which indicates color and a one-hot vector which indicates shape at each pixel of the BEV map.

\begin{figure*}[t]
    \centering
    \includegraphics[width=1.0\textwidth]{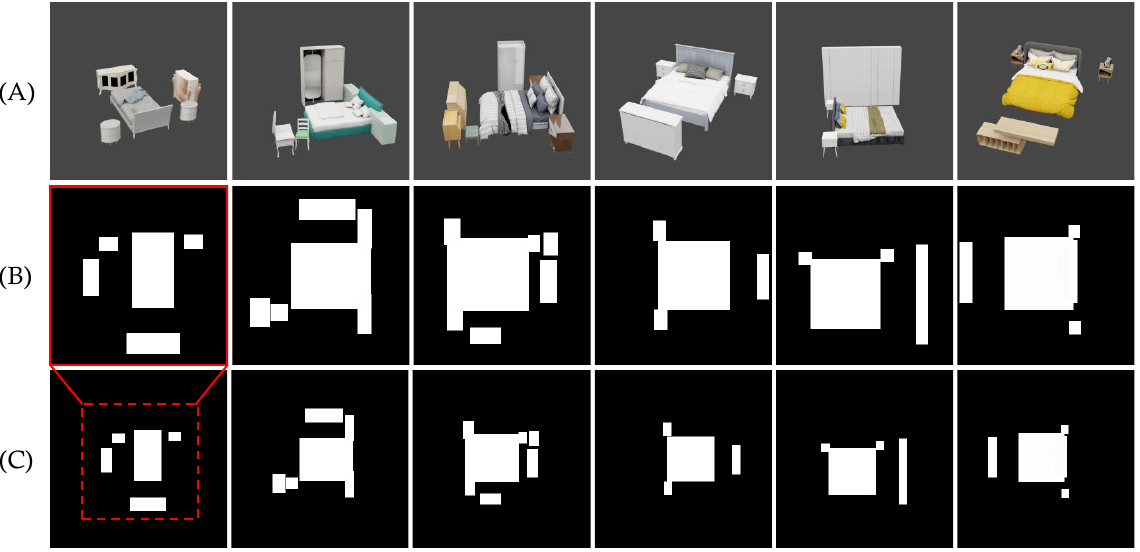}
    \caption{
        \textbf{Sampled front view images and BEV maps on 3D-Front.} Row A shows rendered front view images. Row B and C show corresponding BEV maps without and with broader paddings.
        } 
    \label{fig:supp 3dfront samples}
\end{figure*}

\noindent\textbf{3D-Front.} 3D-Front~\cite{3dfront, 3dfuture} is an indoor scene dataset, which contains different kinds of furniture with fine details.
We use the public script\footnote{https://github.com/DLR-RM/BlenderProc/blob/main/examples} for rendering. 
We filter out objects with abnormal sizes and collect 2535 different scenes in total. For each scene, we render 20 images from different camera poses.
\cref{fig:supp 3dfront samples} shows sampled pairs of rendered images and BEV maps.
Similar to CLEVR, for each scene, we prepare a tight BEV map for the ablation study, and also a broader version for sake of equivariance.
The channel number of the BEV map is one. For each pixel, 0 indicates not occupied by any furniture, while 1 indicates occupied.
We do not include any categorical information in the BEV map. Instead, the generator shall infer such knowledge from size, shape, and relative positions between different objects.

\begin{figure*}[t]
    \centering
    \includegraphics[width=1.0\textwidth]{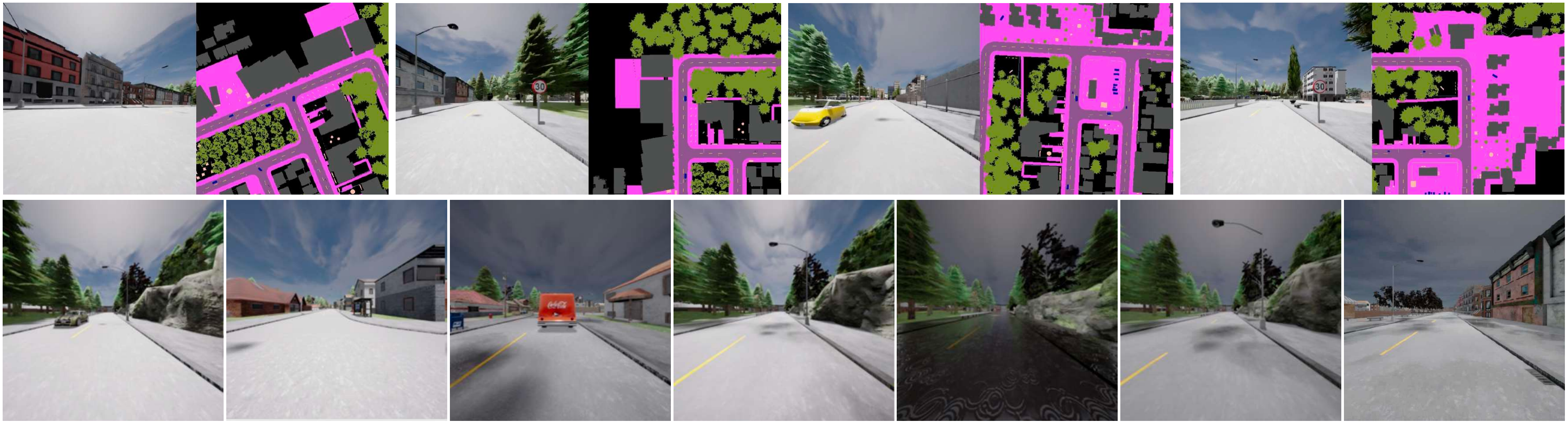}
    \caption{
        \textbf{Sampled front view images and BEV maps on Carla.} The up row shows paired front view images and BEV maps. The bottom row shows diverse weather conditions.
        } 
    \label{fig:supp carla samples}
\end{figure*}

\noindent\textbf{Carla.} Carla~\cite{dosovitskiy2017carla} is a self-driving research simulator that offers a variety of realistic visual patterns, including diverse weather conditions and different types of scenes ranging from rural to urban. In our research, we employ a car equipped with a PID controller to autonomously navigate through the town, capturing images with a front-facing camera. A total of 80K images are collected during the process. The relative camera positions to the car remain fixed for all the images. Additionally, we generate the semantic bird's-eye view (BEV) map following the official primitive guidelines. \cref{fig:supp carla samples}  shows sampled images and BEV maps.

\section{Implementation Details}

We implemented a U-Net architecture for our generator, which consists of four encoders followed by four decoders. Our input is a Fourier feature of shape $256\times256\times256$, which is computed by StyleGAN3's \texttt{SynthesisInput} module. Each encoder downsamples the feature map by a factor of 2 until it reaches a resolution of $16\times16$.

Each encoder in our U-Net architecture includes a downsample layer, a low-pass filter, an SEL module, and two layers of modulated convolutions. The low-pass filter is designed as a finite impulse response (FIR) filter. The kernel size in the modulated convolutions is 3, while it is 1 in the SEL module.
The SEL module takes the similar design as in \cite{wang2022improving}, while we add a low-pass filter after the downsampling operation.
The decoders share a similar architecture design with the encoders, except that there is no low-pass filter in the decoders. This is because the upsampling operation in the decoders does not limit the bandwidth of the signal.

\section{Infinite Generation}

In this section, we make a detailed discussion about how to perform infinite generation over CLEVR and provide more visual examples.

\begin{figure*}[t]
    \centering
    \includegraphics[width=1.0\textwidth]{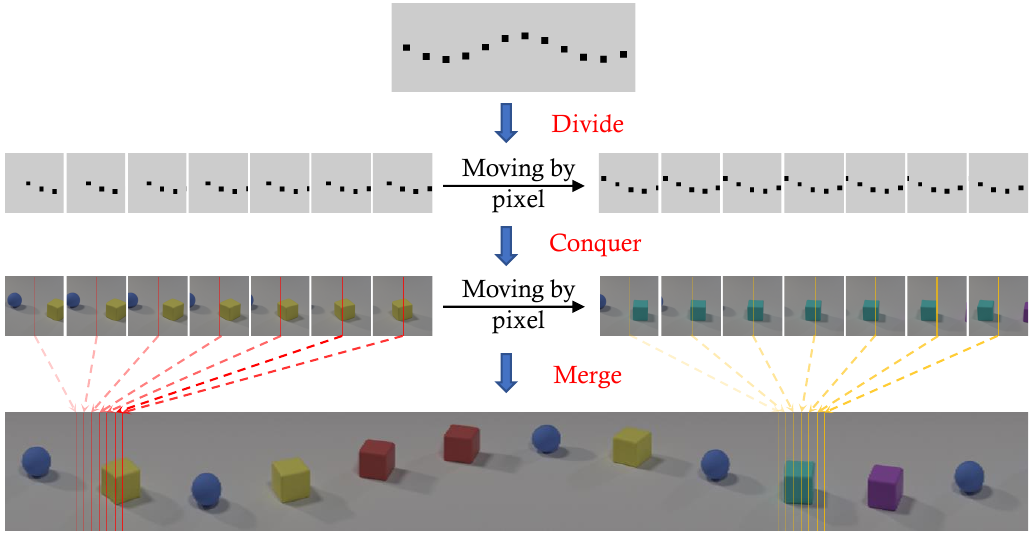}
    \caption{
        \textbf{Illustration of how to perform infinite-scale 3D scene generation.}  
        } 
    \label{fig:supp divide conquer}
\end{figure*}

\begin{figure*}[t]
    \centering
    \includegraphics[width=1.0\textwidth]{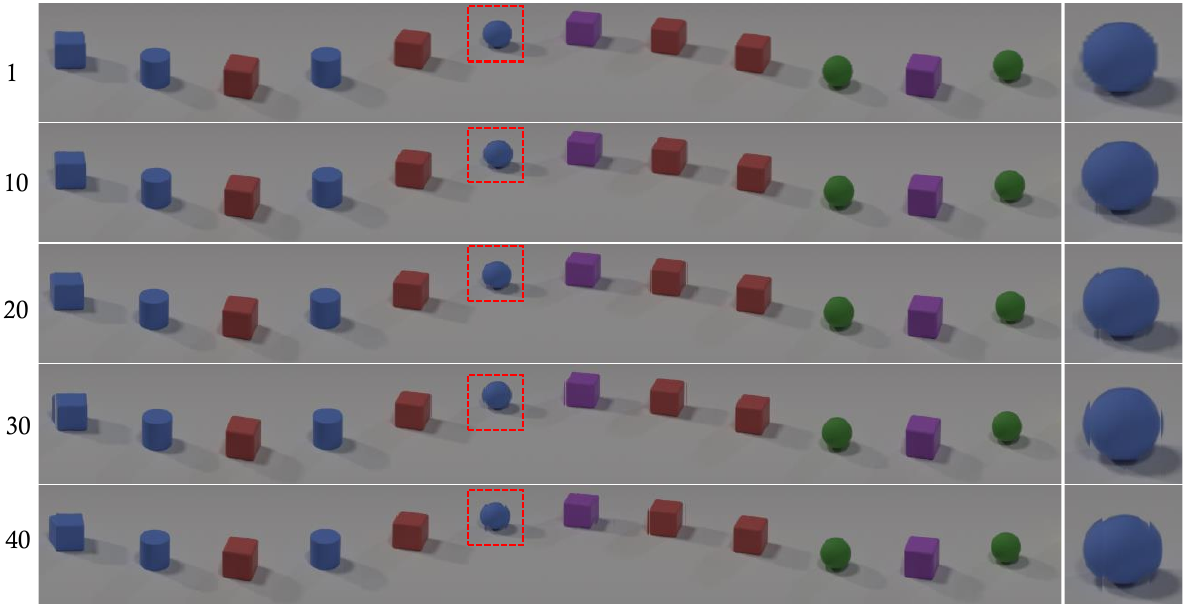}
    \caption{
        \textbf{Synthesized results over different $N_{step}$ choices.}  
        } 
    \label{fig:supp optimize}
\end{figure*}

\noindent\textbf{How to synthesize infinite 3D scene?}
As illustrated by \cref{fig:supp divide conquer}, we generate arbitrary-scale 3D scenes in a \textit{divide-and-conquer} manner.
To generate global scenes, we begin by dividing the global BEV maps into smaller local ones by a sliding window. Using these local BEV maps as input, we generate 3D scenes and obtain multiple first view images.
To form the final global scene, we extract the middle line of pixels from each image and concatenate them together. This process allows us to combine the information from all the local BEV maps and generate a complete representation of the global scene.
It is worth mentioning that, during the \textit{divide} stage, the moving window is shifted pixel by pixel.

Such a design for infinite-scale scene generation places a significant demand on the equivariance property of the generator, as it requires the generator to maintain consistency at a pixel granularity level.
An additional benefit of this approach is that by generating local frames and combining them, we can obtain a traversing video: by simply stacking the generated frames, we can create a video that allows for seamless exploration of the entire scene.
Videos are zipped in the \textit{Supplementary Material}.

If we do not need traversing video, but only want to get a composite image of the global scene, we can optimize the pipeline by increasing the sliding window step size to $N_{step}$. This approach involves collecting $N_{loc}$ consecutive lines of pixels from each synthesized image and concatenating them to form the global view. Leveraging the perspective relationship, we can determine that $N_{loc}$ is equal to $\frac{1}{f_{norm}}\cdot N_{step}$, where $f_{norm}$ represents the normalized focal length.
\cref{fig:supp optimize} shows the results when $N_{step}$ equals 1, 10, 20, 30, 40. 
Serrated artifacts can be observed as $N_{step}$ increases, while $N_{step}=10$ achieves a good balance between efficiency and quality of large-scale 3D scene synthesis.

\begin{figure*}[t]
    \centering
    \includegraphics[width=1.0\textwidth]{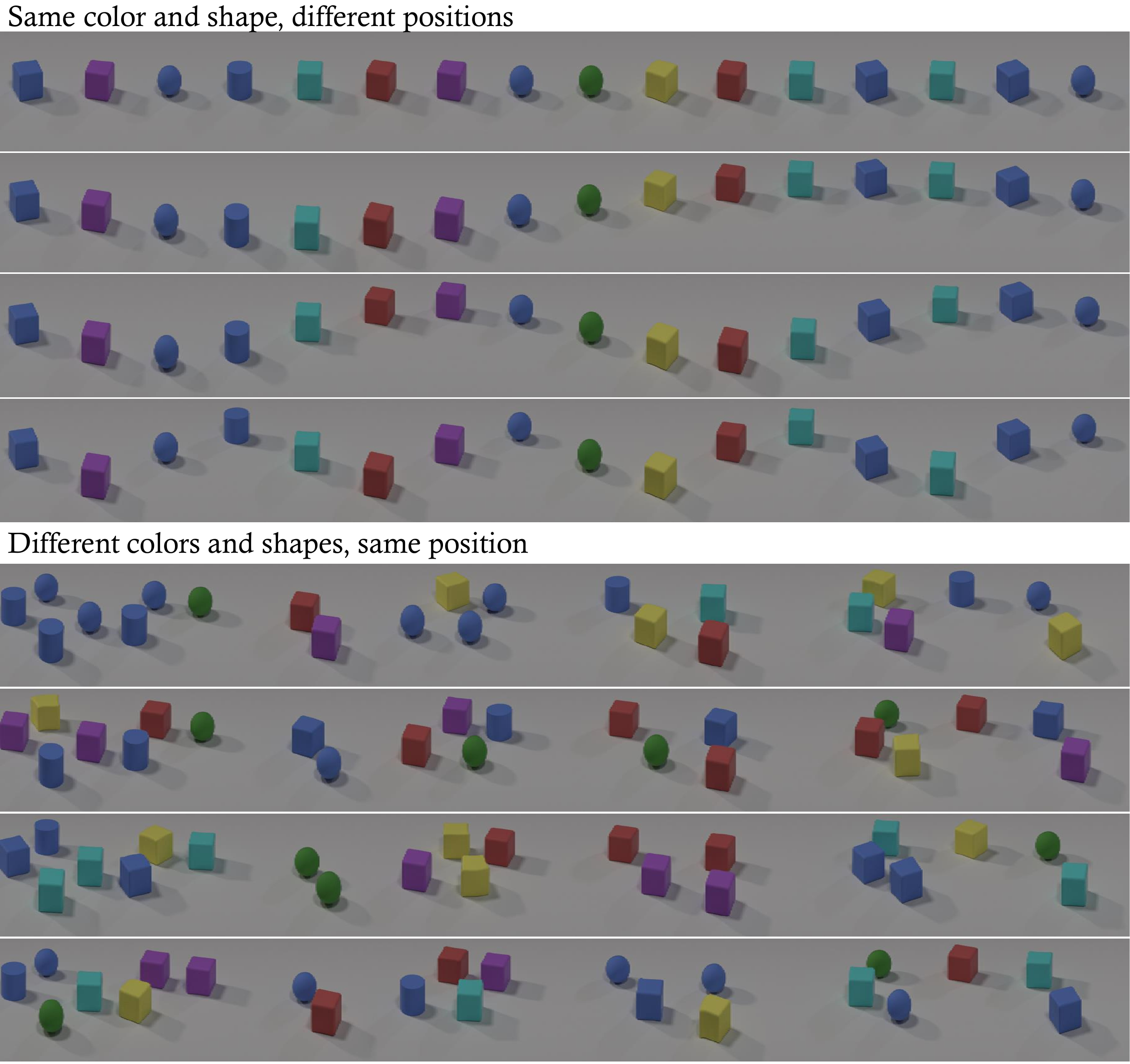}
    \caption{
        \textbf{Synthesized large-scale 3D scene.}  
        } 
    \label{fig:supp infinity}
\end{figure*}

\noindent\textbf{More samples.}
We show more synthesized large scene in \cref{fig:supp infinity}. The corresponding traversing videos could be found at the \textit{Supplementary Material}.

\end{document}